%% file: main.tex
\pdfoutput=1 % Signal to arXiv to compile with PDFLaTex
\documentclass[10pt,twocolumn,letterpaper]{article}

\usepackage{cvpr}              % To produce the CAMERA-READY version
\usepackage{graphicx}
\usepackage{amsmath}
\usepackage{amssymb}
\usepackage{booktabs}

\makeatletter
\@namedef{ver@everyshi.sty}{}
\makeatother

\newcommand{\OURS}{ZippyPoint}

\usepackage{tabularx}
\usepackage{multirow} 
\usepackage[dvipsnames]{xcolor}
\newcommand{\parsection}[1]{\vspace{1mm}\noindent\textbf{#1 }}
\usepackage{witharrows}
\usepackage{color, colortbl}
\definecolor{Gray}{gray}{0.90}
\usepackage{dblfloatfix}

\usepackage[acronym]{glossaries}

\newacronym{DNNs}{DNNs}{Deep Neural Networks}
\newacronym{SLAM}{SLAM}{Simultaneous Localization and Mapping}
\newacronym{AR}{AR}{Augmented Reality}
\newacronym{VisLoc}{VisLoc}{Visual Localization}
\newacronym{SfM}{SfM}{Structure-from-Motion}
\newacronym{QNN}{QNN}{Quantized Neural Network}
\newacronym{MP}{MP}{Mixed-Precision}
\newacronym{FP}{FP}{Floating-Point}
\newacronym{Int}{Int}{Integer}
\newacronym{Bin}{Bin}{Binary}
\newacronym{FPS}{FPS}{Frames Per Seconds}
\newacronym{AUC}{AUC}{Area Under the Curve}
\newacronym{Err}{Err}{pose error}
\newacronym{VCRE}{VCRE}{Virtual Correspondence
Reprojection Error}

\newacronym{Bin-R}{Bin-R}{Binary with a high-precision Residual}
\newacronym{Bin.Norm}{Bin.Norm}{Binary Normalization}

\NewExpandableDocumentCommand{\gcmidrule}{ O{} D(){} m }{%
    \arrayrulecolor{lightgray}%
    \cmidrule{#3}%
    \arrayrulecolor{black}%
}

\usepackage[pagebackref,breaklinks,colorlinks]{hyperref}

\usepackage[capitalize]{cleveref}
\crefname{section}{Sec.}{Secs.}
\Crefname{section}{Section}{Sections}
\Crefname{table}{Table}{Tables}
\crefname{table}{Tab.}{Tabs.}

\def\cvprPaperID{3055} % *** Enter the CVPR Paper ID here
\def\confName{CVPR}
\def\confYear{2023}

\begin{document}

\title{\OURS: Fast Interest Point Detection, Description, and Matching through Mixed Precision Discretization}

\author{Menelaos Kanakis$^{*,1}$\quad
        Simon Maurer$^{*,1}$\quad
        Matteo Spallanzani$^{1}$\quad
        Ajad Chhatkuli$^{1}$\quad
	Luc Van Gool$^{1,2}$ \vspace{2mm} \\
$^1$ETH Z\"urich \quad $^2$KU Leuven
}
\maketitle

\input{text/abstract}

{\let\thefootnote\relax\footnote{{$^*$M. Kanakis and S. Maurer contributed equally to this work.}}}

\input{text/intro}
\input{text/relwork}
\input{text/method}
\input{text/exp}

\input{text/conclusion}

\begin{center}
	\textbf{\Large Supplementary Material}
\end{center}

\input{text_supp/visloc_additional.tex}

\input{text_supp/mapfreeloc.tex}

\begin{figure}[b]
\vspace{8.5in}
\end{figure}

\newpage

%%%%%%%%% REFERENCES
{\small
\bibliographystyle{ieee_fullname}
\bibliography{main}
}

\end{document}

%% file: text/abstract.tex
%%%%%%%%% ABSTRACT
\begin{abstract}
    Efficient detection and description of geometric regions in images is a prerequisite in visual systems for localization and mapping.
    Such systems still rely on traditional hand-crafted methods for efficient generation of lightweight descriptors, a common limitation of the more powerful neural network models that come with high compute and specific hardware requirements.
    In this paper, we focus on the adaptations required by detection and description neural networks to enable their use in computationally limited platforms such as robots, mobile, and augmented reality devices. 
    To that end, we investigate and adapt network quantization techniques to accelerate inference and enable its use on compute limited platforms. 
    In addition, we revisit common practices in descriptor quantization and propose the use of a binary descriptor normalization layer, enabling the generation of distinctive binary descriptors with a constant number of ones.
    \OURS{}, our efficient quantized network with binary descriptors, improves the network runtime speed, the descriptor matching speed, and the 3D model size, by at least an order of magnitude when compared to full-precision counterparts.
    These improvements come at a minor performance degradation as evaluated on the tasks of homography estimation, visual localization, and map-free visual relocalization.
    Code and models are available at \url{https://github.com/menelaoskanakis/ZippyPoint}.
\end{abstract}

%% file: text/intro.tex
\begin{figure}[t!]
\vspace{-0.1in}
 \centering
  \includegraphics[width=\linewidth]{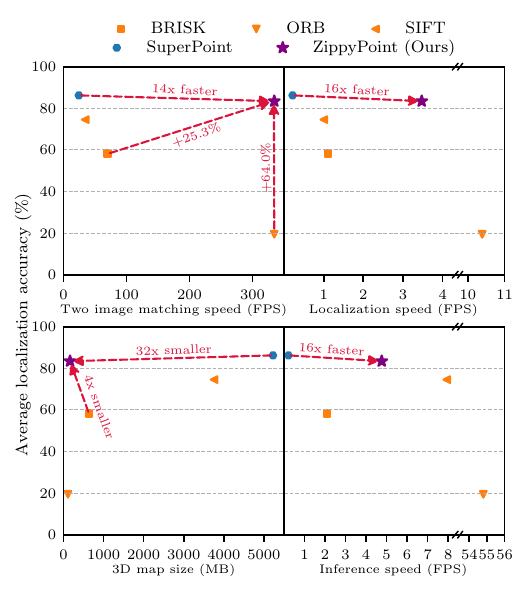}
  \vspace{-0.3in}
\caption{Learned detection and description methods, \eg SuperPoint~\cite{detone2018superpoint}, significantly outperform hand-crafted methods, in orange, on challenging day-night scenarios~\cite{sattler2018benchmarking,sattler2012image}.
This, however, comes at the cost of slower image matching (top left), slower keypoint detection and description (bottom right), larger 3D models (bottom left), and therefore slow localization within a 3D map (top right). 
Speeds are reported in FPS on a CPU.
In this paper we present \OURS{}, a learned detection and description network that improves the above limitations by at least an order of magnitude while providing competitive performance, enabling its use on-board computationally limited platforms.}
  \label{fig:teaser}
 \vspace{-0.3in}
\end{figure}

\section{Introduction}
\label{sec:intro}
The detection and description of geometric regions in images, such as salient points or lines, is one of the fundamental components in visual localization and mapping pipelines -- essential prerequisites for \gls{AR} and robotic applications. 
Achieving such detection and description efficiently with handcrafted algorithms~\cite{rublee2011orb,leutenegger2011brisk} has produced successful robot localization methods~\cite{mur2015orb,mur2017orb,endres2012evaluation,galvez2012bags,leutenegger2015keyframe}.
On the other hand, \gls{DNNs} have significantly advanced the representational capability of descriptors by learning on large scale natural images~\cite{detone2018superpoint}, using deeper networks~\cite{dusmanu2019d2}, or introducing new modules to learn feature matching~\cite{sarlin2020superglue}.
However, these advances often come at the cost of more expensive models with slow run times and large memory requirements for representation storage, making them unsuitable for computationally limited platforms.
While the demand for real-time applications such as robotics and \gls{AR} is increasing, efficient DNN methods that can operate in real-time on computationally limited platforms have received surprisingly little attention.

A key component for the successful deployment of mobile robots in large-scale applications is the real-time extraction of binary descriptors.
This not only enables efficient storage of the detected representation, e.g. the map in \gls{SLAM} or \gls{SfM} pipelines, but also accelerated descriptor matching.
In particular, matching computations in localization scale non-linearly with the number of images or map size.
Therefore, improved two-view matching speed can translate to very high gains in real applications.
Fast and light weight descriptor methods include BRISK~\cite{leutenegger2011brisk}, BRIEF~\cite{calonder2010brief} and ORB~\cite{rublee2011orb}, however, their matching capability is often inferior to standard hand-crafted features such as SIFT~\cite{lowe2004distinctive} and SURF~\cite{bay2006surf}, as presented by Heinly~J.~\textit{et~al.}~\cite{heinly2012comparative}.
In challenging scenarios, however, hand-crafted feature extractors are outperformed significantly by learned representations~\cite{sattler2018benchmarking}.
While the performance gains of learned methods are highly desired, embedded platforms are limited in storage, memory, providing limited or no support for \gls{FP} arithmetic, thus limiting the use of learned methods. 

Motivated by the desire for improving the performance of feature points on low-compute platforms, we explore DNN quantization to enable the real-time generation of learned descriptors under such challenging constraints.
However, the quantization of a DNN is not as straightforward as selecting the discretization level of convolutional layers.
Quantized \gls{DNNs} often require different levels of discretized precision for different layers \cite{rastegari2016xnor,liu2018bi}.
Operations such as max-pooling favour saturation regimes~\cite{rastegari2016xnor}, while average pooling is affected by the required rounding and truncation operations.
Moreover, prior works often focus on image-level classification tasks, with findings that do not necessarily transfer to new tasks~\cite{bethge2019simplicity}.
To render the search for a \gls{QNN} tractable, we propose a \emph{layer partitioning and traversal} strategy, significantly reducing the architecture search complexity.
While most research considers homogeneous quantization precision across all layers~\cite{nagel2019data}, we find \gls{MP} quantization yields superior performance.
In addition, we find that replacing standard pooling operations with learned alternatives can further improve \gls{QNN} performance.

Besides the need for real-time inference, \gls{DNNs} need to additionally generate binary local descriptors for storage efficiency and fast feature matching.
This adds further challenges as the discretization of the output layer draws less precise boundaries in the feature domain~\cite{leonardi2020analytical}, making the network optimization more challenging.
Furthermore, prior works focus on global feature description and present findings that do not trivially transfer to our task~\cite{shen2018unsupervised,lai2015simultaneous}.
To this extent, we introduce a \gls{Bin.Norm} layer that constrains the representation to a constant pre-defined number of ones.
\gls{Bin.Norm} is therefore analogous to the $L_2$ normalization, a staple and key component in \gls{FP} metric learning~\cite{musgrave2020metric}.

In summary, our contributions are:
\begin{itemize}
    \item We propose a heuristic algorithm, named \textit{layer partitioning and traversal strategy}, to investigate the topological changes required for the quantization of a state-of-the-art detection and description network. 
    We find that with a \gls{MP} quantization architecture, and by replacing max-pooling with a learned alternative, we can achieve a speed-up by an order of magnitude with minor performance degradation.
    Our analysis reveals that the common \gls{QNN} practices can be sub-optimal.
    \item We propose the use of a normalization layer for the end-to-end optimization of binary descriptors.
    Incorporating the \gls{Bin.Norm} layer yields consistent improvements when compared to the common practices for descriptor binarization.
    \item We provide a detailed analysis of \OURS{}, our proposed \gls{QNN} with binary descriptors, on the task of homography estimation. We further demonstrate the generality of \OURS{} on the challenging applications of \gls{VisLoc} and Map-Free Visual Relocalization. 
    \OURS{} consistently outperforms all real-time alternatives and yields comparable performance to a full precision counterpart while addressing its known limitations, illustrated in Fig.~\ref{fig:teaser}.
\end{itemize}

%% file: text/relwork.tex
\section{Related Work}
\label{sec:relwork}

\parsection{Hand-crafted feature extractors.} The design of hand-crafted sparse feature extractors such as SIFT~\cite{lowe2004distinctive} and SURF~\cite{bay2006surf} has been undoubtedly very successful in practice, still widely used in applications such as SfM~\cite{schonberger2016structure}.
However, the time needed for detection and descriptor extraction, coupled with their \gls{FP} representation, limits them from being used on compute-limited platforms, such as light-weight unmanned aerial vehicles.
Motivated by this limitation, methods like BRISK~\cite{leutenegger2011brisk}, BRIEF~\cite{calonder2010brief}, and ORB~\cite{rublee2011orb}, aimed to provide compact features targeted for real-time applications~\cite{mur2015orb,mur2017orb,leutenegger2015keyframe}.
While fast and lightweight, they lack the representational strength to perform well under a wide variety of viewing conditions such as large viewpoint changes~\cite{heinly2012comparative,schonberger2016structure} or times of day and year~\cite{sattler2018benchmarking}.

\begin{figure*}[t]
 \centering
  \subfloat[Detection and Description architecture]{\label{fig:teaser_rt}
      \includegraphics[width=.42\textwidth]{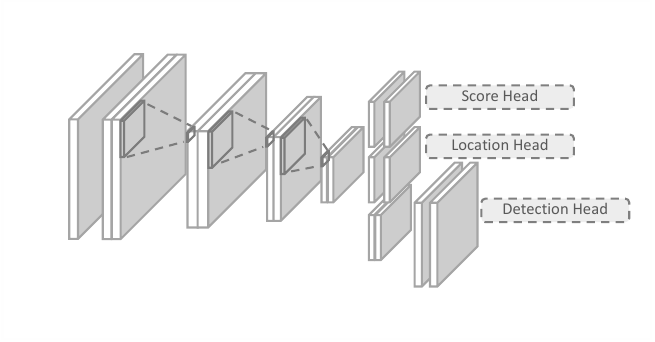}}
% \hfill
      \subfloat[The layer partitioning and traversal strategy]{\label{fig:teaser_match}
      \includegraphics[width=.42\textwidth]{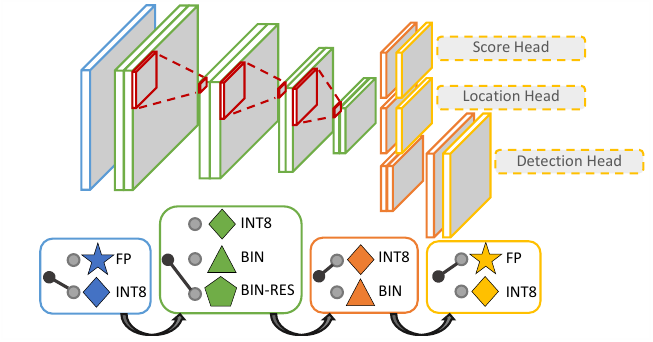}}
\caption{(a) Starting from \cite{tang2019neural}, we partition the operations in macro-Blocks, depicted in (b)  with different colors. From the first upstream macro-Block, in blue, we identify the optimal quantization setting that maintains functional performance while improving the network's throughput. We then traverse to the next block, green, and repeat. The strategy is complete once we have reached the most downstream network layer, the prediction heads.}
  \label{fig:architecture}
 \vspace{-0.2in}
\end{figure*}

\parsection{Learned feature extractors.} Advances in \gls{DNNs} have enabled the learning of robust, (pseudo-)invariant, and highly descriptive image features, pushing the boundaries of what was previously possible through hand-crafted methods. 
% Recent efforts have also brought the aforementioned advantages of learning methods to local image point features. 
While hand-crafted local features~\cite{lowe2004distinctive,bay2006surf,leutenegger2011brisk,calonder2010brief,rublee2011orb,tola2009daisy} have not evolved much, systematic incremental progress can be seen in the learned local features~\cite{choy2016universal,detone2018superpoint,fathy2018hierarchical,ono2018lf,revaud2019r2d2,tang2019neural,dusmanu2019d2}. 
Improvements have been achieved using contrastive learning~\cite{choy2016universal}, self-supervised learning~\cite{detone2018superpoint}, improved architectures~\cite{revaud2019r2d2,tang2019neural} and outlier rejection~\cite{tang2019neural}, to name a few approaches. 
Nevertheless, time and memory inefficiency remain major drawbacks of the learned methods.

In the same vein, large scale descriptor matching calls for light-weight representations.
Binary descriptors enable efficient matching with moderate performance drops while significantly decreasing the storage requirements. 
Yet, the existing literature on binary representations focuses on image retrieval~\cite{lin2016learning,shen2018unsupervised,lai2015simultaneous,shen2015supervised,wang2017survey,norouzi2012hamming}, neglecting the detection and description of local features. 
For descriptor binarization, \cite{shen2018unsupervised,shen2015supervised,lin2016learning} use multistage optimization procedures. 
More similar to our work, \cite{norouzi2012hamming} defines a differentiable objective for the hamming distance, \cite{lai2015simultaneous} uses sigmoids to soften the optimization objective, while \cite{tang2019gcnv2} rely on a hard sign function and gradient approximations.
In the same spirit, we also optimize the network in a single optimization step.
However, we argue that the lack of normalization layer in these methods, a staple in metric learning~\cite{musgrave2020metric}, greatly hinders the descriptor performance.
To address this limitation, we propose a normalization layer for binary descriptors.
\gls{Bin.Norm} provides a more stable optimization process, avoids mode collapse and enables end-to-end optimization without requiring the use of gradient approximations or multiple optimization stages.

\parsection{Efficient Neural Networks.} 
Several solutions have been proposed to deploy neural networks in constrained scenarios.
These solutions can be partitioned in topological optimizations, aiming at increasing accuracy-per-operation or accuracy-per-parameter~\cite{he2015convolutional,howard2019searching,blatter2023efficient}, software optimizations such as tensor decomposition and parameter pruning~\cite{zhang2015accelerating,obukhov2020t,obukhov2021spectral}, and hardware-aware optimizations~\cite{rastegari2016xnor}.

Amongst hardware-aware optimizations, quantization plays a central role~\cite{rastegari2016xnor,liu2018bi}.
By replacing \gls{FP} with \gls{Int} operands, a \gls{QNN} can reduce its storage and memory requirements with respect to an equivalent DNN.
In addition, complex \gls{FP} arithmetics can be replaced by simpler \gls{Int} arithmetics.
Due to these properties, \gls{QNN}s can be executed at higher throughput (number of operations per cycle) and arithmetic intensity (number of arithmetic operations per memory transaction)~\cite{jacob2018quantization}.
When operand precision is extremely low, e.g.\ \gls{Bin}, standard instruction set architectures can be exploited to increase these metrics even further~\cite{rastegari2016xnor}.
Unlike mainframes and workstations, embedded platforms have limited storage and memory, limited or no support for \gls{FP} arithmetics, and are optimized to execute SIMD (Single Instruction, Multiple Data) \gls{Int} arithmetics.
These considerations make \gls{QNN}s an ideal fit for embedded applications, such as robots and mobile devices.

However, \gls{QNN}s have limited representational capacity compared to their \gls{FP} counterparts.
Specifically, linear operations using discretized weights draw less precise boundaries in their input domains.
In addition, discretized activation functions lose injectivity with respect to their \gls{FP} counterparts, making quantization a lossy process~\cite{leonardi2020analytical}.
To strike a balance between throughput and performance, practitioners require to identify a single \gls{Int} precision~\cite{nagel2019data}, or alternative linear layers~\cite{liu2018bi}, that achieve the desired performance.
These design choices are applied homogeneously across the entire network.
We instead hypothesize that a single set of hyperparameters across the entire network can be suboptimal.
We, therefore, investigate the use of heterogeneous layers throughout the network, e.g.\ different \gls{Int} precision at different depths of the network, made possible through the proposed layer partitioning and traversal strategy.

%% file: text/method.tex
\section{Mixed Precision Discretization}
\label{sec:method}

Efficiently identifying salient points in images and encoding them with lightweight descriptors is key to enabling real-time applications such as robot localization.
In this paper we explore the efficacy of learning-based descriptor methods under two constraints: minimizing run-time latency and using binary descriptors for accelerating keypoint matching and efficient storage.
In Sec.~\ref{subsec:architecture} we introduce the baseline architecture we initiate our investigation from. 
In Sec.~\ref{subsec:network_structure} we propose a strategy to explore structural changes to the network's topology. 
In Sec.~\ref{subsec:BNL} we introduce a standard formulation of metric learning, which we use to then define our \gls{Bin.Norm} descriptor layer.

\subsection{Baseline Architecture}
\label{subsec:architecture}

We initiate our investigation from the state-of-the-art KP2D~\cite{tang2019neural} network, which exploits outlier filtering to improve detections.
The KP2D model maps an input image $\mathbf{I} \in \mathbb{R}^{H\times W\times3}$ to keypoints $\mathbf{p} \in \mathbb{R}^{N \times 2}$, descriptors $\mathbf{x} \in \mathbb{R}^{N\times M}$, and keypoint scores $\mathbf{s} \in \mathbb{R}^{N}$, where $N$ represents the total number of keypoints extracted and $M$ the descriptor size.
The model is comprised of an encoder with 4 VGG-style blocks~\cite{simonyan2014very}, followed by a three-headed decoder for keypoints, keypoint scores, and descriptors.
% The distinction between encoder and decoder is decided by the reduction of or increase in the spatial resolution of the input.
The encoder is comprised of 8 convolutional operations, the keypoint and keypoint score branches of 2, and the descriptor branch of 4.
All convolutional operations, except for the final layers, are followed by batch normalization and leaky ReLUs~\cite{maas2013rectifier}.
The model is optimized through self-supervision by enforcing consistency in predictions between a source image $\mathbf{I}_s$ and a target image $\mathbf{I}_t = \text{H}(\mathbf{I}_s)$, related through a known homography transformation $\mathbf{H}$ and its warping function H.

We chose KP2D as the starting point for our investigation due to its standard architecture design choices:
a VGG style encoder~\cite{detone2018superpoint,christiansen2019unsuperpoint,dusmanu2019d2,revaud2019r2d2}, encoder-decoder structure~\cite{detone2018superpoint,christiansen2019unsuperpoint}, and the detection and description paradigm~\cite{detone2018superpoint,christiansen2019unsuperpoint,dusmanu2019d2,revaud2019r2d2}.
Therefore, we expect that the investigated quantization strategy can transfer to other similar models, such as the ones listed above. 

\subsection{Network Quantization}
\label{subsec:network_structure}

For the quantization of a convolutional layer, several design choices are required.
These include weight precision, feature precision, and whether to use a high precision residual.
When considering independently each layer of a DNN, it leads to a combinatorially large search grid, rendering an exhaustive search of the ideal quantization policy prohibitive.

To simplify the search space, we propose the \emph{layer partitioning and traversal strategy}, depicted in Fig.~\ref{fig:architecture}.
First, we partition the operations of our target architecture into macro-blocks.
For each macro-block, we define a collection of candidate quantized configurations.
We then traverse through the macro-blocks and identify the optimal configuration for each, one at a time.
This heuristic algorithm terminates once we have reached the most downstream network layer, the prediction heads.
Note that, while we maintain the macro-block configurations the same once selected, the architecture is always optimized end-to-end.
This strategy reduces the search complexity from combinatorial (the product of the number of configurations for each macro-block) to linear (the sum of the number of configurations for each macro-block).
In addition, it ensures that when a macro-block is optimized on features with given representation capabilities, it will not degrade due to optimization of a different macro-block upstream.
We detail our choice of macro-blocks and their configurations in the experiments section.

\subsection{Binary Learned Descriptors}
\label{subsec:BNL}

\parsection{Preliminaries.}
When describing an image or a local region, the learned mapping aims to project a set of data points to an embedding space, where similar data are close together and dissimilar data are far apart.
A fundamental component to the success of learned descriptors is the advancement of contrastive losses~\cite{hadsell2006dimensionality,weinberger2009distance,deng2019arcface}.
To ensure stable optimization and avoiding mode collapse, descriptors are often normalized~\cite{musgrave2020metric}.
A common selection is $L_2$ normalization, defined as
\begin{equation}
\mathbf{y}=\frac{1}{||\mathbf{x}||_2}\mathbf{x}.
\label{eq:l2_norm}
\end{equation}
While the solution, and hence gradients, can be expressed in closed-form, it assumes \gls{FP} representation spaces $\mathbf{x}$ and $\mathbf{y}$.
However, \gls{Bin} descriptors can only take discrete values \{0,1\}.
In search for a normalization layer applicable for \gls{Bin} descriptors, we instead view and rewrite Eq.~\eqref{eq:l2_norm} as the generalized optimization objective
\begin{equation}
\begin{split}
\mathbf{y} = \underset{\mathbf{z} \in \mathbb{R}^M}{arg\min} \quad d(\mathbf{z}; \mathbf{x})\\
\text{subject to} \quad constr(\mathbf{z}),
\label{eq:norm_objective}
\end{split}
\end{equation}
where we search for the vector $\mathbf{z}$ that minimizes a distance function to $\mathbf{x}$ under a normalization constraint $constr(\mathbf{z})$.
Eq.~\eqref{eq:norm_objective} is therefore equivalent to Eq.~\eqref{eq:l2_norm} when $d(\mathbf{z}; \mathbf{x}) = \frac{1}{2}||\mathbf{z}-\mathbf{x}||^2_2$ and $constr(\mathbf{z})$ is $||\mathbf{z}||_2 = 1$.
This enables the definition of optimization objectives that can be utilized where $L_2$ normalization does not provide the required behaviour. 
While constrained optimization problems are not differentiable, their use in \gls{DNNs} is made possible through advances in deep declarative networks~\cite{gould2021deep}.

\parsection{Normalization for Binary Descriptors.}
We hypothesize that normalization for binary descriptors is equivalent to having a constant number of ones in each descriptor.
To this end, we take inspiration from multi-class classification problems and view the \gls{Bin.Norm} as a projection of the descriptors living in an M-dimensional hypercube on a $k$-dimensional polytope~\cite{amos2019limited}. 
In other words, an $M$-dimensional descriptor has entries that sum to $k$.
This trivially yields the constraint from Eq.~\eqref{eq:norm_objective} to $constr(\mathbf{z}) = \mathbf{1}^\top \mathbf{z} = k$, where $\mathbf{1}$ is a vector of 1s of the same dimension as $\mathbf{z}$.

We define the new optimization objective as
\begin{equation}
\begin{split}
\mathbf{y} = \underset{\mathbf{z}\in [0,1]^M}{arg\min}  -\mathbf{x}^\top \mathbf{z} -H(\mathbf{z})\\
\text{subject to} \quad \mathbf{1}^\top \mathbf{z} = k
\label{eq:bin_norm}
\end{split}
\end{equation}
where $H(\mathbf{z})$ is the binary entropy function applied on the vector $\mathbf{z}$ for entropy based regularization.

To optimize the objective, we introduce a dual variable $\nu\in\mathbb{R}$ for the constraint of Eq.~\eqref{eq:bin_norm}.
The Lagrangian then becomes
\begin{equation}
-\mathbf{x}^T\mathbf{z} - H(\mathbf{z}) + \nu(k-\mathbf{1}^\top \mathbf{z}).
\label{eq:langrangian}
\end{equation}
Differentiating with respect to $\mathbf{z}$, and solving for first-order optimality gives
\begin{equation}
-\mathbf{x}+\log \frac{\mathbf{z}^*}{\mathbf{1}-\mathbf{z}^*} - \nu^* = 0,
\label{eq:optimal_1}
\end{equation}
that yields
\begin{equation}
\begin{split}
\mathbf{y} \approx \mathbf{z}^*=\sigma(\mathbf{x}+\nu^*),
\label{eq:optimal}
\end{split}
\end{equation}
where $\sigma$ denotes the logistic function.
We identify the optimal $\nu^*$ by using the bracketing method of \cite{amos2019limited}, efficiently implemented for use on GPUs, and backpropagate using~\cite{amos2017optnet}.

The selection of the optimization objective in Eq.~\eqref{eq:bin_norm} is two-fold. The entropy regularizer helps prevent sparsity in the gradients of the projection.
In addition, the forward pass in Eq.~\eqref{eq:optimal} can be seen as an adaptive sigmoid that ensures the descriptor entries sum up to a specific value. 
This enables the direct comparison with the common practice of approximating binary entries using the sigmoid function.

At inference, the descriptor optimization strategy in
Eq.~\ref{eq:bin_norm} is replaced by a thresholding function that sets the top-k logits of each descriptor to one, for faster processing.
We set the top-64 to ones, however, we found that both smaller and larger numbers yield a comparable performance.

\begin{table*}%[ht]
  \center
    \caption{Results from the layer partitioning and traversal strategy. The final model, in bold, performs comparably to the baseline while running an order of magnitude faster. Green highlights the configuration used in the next stage.}
    \vspace{-0.1in}

    \label{table:layer_part}
      \resizebox{\linewidth}{!}{%
      \scriptsize
  \begin{tabularx}{\linewidth}{@{}ccccccXXXXXXc@{}}
    \toprule
    &&&&&& Repeat. $\uparrow$ & Loc. $\downarrow$ & Cor-1 $\uparrow$ & Cor-3 $\uparrow$ & Cor-5 $\uparrow$ & M.Score $\uparrow$ & FPS $\uparrow$ \\
    \cmidrule{7-13}
	&\multicolumn{5}{l}{{KP2D \cite{tang2019neural}}} & 0.686 & 0.890 & 0.591 & 0.867 & 0.912 & 0.544 & 2.5 \\
 \hspace*{0.7cm}\multirow{5}{*}{\reflectbox{$\WithArrowsOptions{displaystyle,tikz=ForestGreen}\begin{WithArrows} & \Arrow{}[jump=3] \\& \\& \\& \\& \end{WithArrows}$}\vspace{-0.5cm}}\hspace*{-0.7cm}
&\multicolumn{5}{l}{{\textcolor{ForestGreen}{Baseline (ours)}}} & 0.649 $\pm$ 0.004 & 0.792 $\pm$ 0.015 & 0.566 $\pm$ 0.015 & 0.880 $\pm$ 0.011 & 0.925 $\pm$ 0.007 & 0.571 $\pm$ 0.004 & 3.6\\
     \cmidrule{2-13}
      
    &\multicolumn{5}{c}{{\textbf{\underline{Encoder Convolutions}}}}\\
    &\multicolumn{2}{c}{First conv} & \multicolumn{3}{c}{Remaining convs}\\
    &\scriptsize{FP} & \scriptsize{Int8}  & \scriptsize{Int8} & \scriptsize{Bin} & \scriptsize{Bin-R}\\
% 	FP init + int8 convs
	&\pmb{\checkmark}&&\pmb{\checkmark}&& & 0.651 $\pm$ 0.004 & 0.840 $\pm$ 0.076 & 0.528 $\pm$ 0.019 & 0.867 $\pm$ 0.017 & 0.922 $\pm$ 0.011 & 0.574 $\pm$ 0.004 & 10.6\\
% 	Int8 init + int8 convs
	 &\hspace*{1cm}\multirow{2}{*}{\reflectbox{$\WithArrowsOptions{displaystyle,tikz=ForestGreen}\begin{WithArrows} & \Arrow{}[jump=1] \\&  \end{WithArrows}$}\vspace{-0.2cm}}\hspace*{-1cm}
&\textcolor{ForestGreen}{\pmb{\checkmark}}&\textcolor{ForestGreen}{\pmb{\checkmark}}&& & 0.657 $\pm$ 0.003 & 0.825 $\pm$ 0.015 & 0.548 $\pm$ 0.003 & 0.866 $\pm$ 0.004 & 0.925 $\pm$ 0.010 & 0.577 $\pm$ 0.001 & 13.8\\
	\gcmidrule{2-13}

% 	Binary convs 
&&\pmb{\checkmark}&&\pmb{\checkmark}& & 0.561 $\pm$ 0.036 & 1.120 $\pm$ 0.061 & 0.281 $\pm$ 	0.132 & 0.401 $\pm$ 0.022 & 0.449 $\pm$ 0.080 & 0.247 $\pm$ 0.096 & 15.2\\
% 	Binary residual convs 
\hspace*{0.7cm}\multirow{4}{*}{\reflectbox{$\WithArrowsOptions{displaystyle,tikz=ForestGreen}\begin{WithArrows} & \Arrow{}[jump=2] \\&  \\& \\& \\& \end{WithArrows}$}\vspace{-1cm}}\hspace*{-0.7cm} &&\textcolor{ForestGreen}{\pmb{\checkmark}}&&&\textcolor{ForestGreen}{\pmb{\checkmark}} & 0.653 $\pm$ 0.005 & 1.040 $\pm$ 0.018 & 0.401 $\pm$ 0.034 & 0.811 $\pm$ 0.011 & 0.890 $\pm$ 0.007 & 0.563 $\pm$ 0.006 & 14.5\\
	\cmidrule{2-13}
    
 &   \multicolumn{5}{c}{{\textbf{\underline{Spatial Reduction}}}}\\
  &  \scriptsize{Max} & \scriptsize{Aver.} & \scriptsize{Sub.S.} & \scriptsize{Learn} & \scriptsize{E.Learn} &\\
% 	Max Pool && 
&	\pmb{\checkmark}&&&& & 0.653 $\pm$ 0.005 & 1.040 $\pm$ 0.018 & 0.401 $\pm$ 0.034 & 0.811 $\pm$ 0.011 & 0.890 $\pm$ 0.007 & 0.563 $\pm$ 0.006 & 14.5\\
% 	Average Pool && 
&&\pmb{\checkmark}&&& & 0.656 $\pm$ 0.003 & 1.068  $\pm$ 0.033 & 0.354 $\pm$ 0.038 & 0.788 $\pm$ 0.027 & 0.873 $\pm$ 0.011 & 0.558 $\pm$ 0.015 & 13.9\\
% 	Subsample && 
&	&&\pmb{\checkmark}&& & 0.640 $\pm$ 0.022 & 1.128  $\pm$ 0.053 & 0.362 $\pm$ 0.024 & 0.783 $\pm$ 0.003 & 0.881 $\pm$ 0.007 & 0.537 $\pm$ 0.015 & 14.4\\
% 	\MK{IPS - Think of a better name} && 
&&&&\pmb{\checkmark}& & 0.648 $\pm$ 0.009 & 0.890 $\pm$ 0.066 & 0.517 $\pm$ 0.029 & 0.848 $\pm$ 0.009 & 0.916 $\pm$ 0.003 & 0.571 $\pm$ 0.006 & 14.2\\
% 	\MK{IPS repositined - Think of a better name} 
&	\hspace*{1cm}\multirow{4}{*}{\reflectbox{$\WithArrowsOptions{displaystyle,tikz=ForestGreen}\begin{WithArrows} & \Arrow{}[jump=3] \\&  \\&  \\& \\& \\& \end{WithArrows}$}\vspace{-1.2cm}}\hspace*{-0.7cm}  &&&&\textcolor{ForestGreen}{\pmb{\checkmark}} & 0.656 $\pm$ 0.002 & 0.943 $\pm$ 0.034 & 0.491  $\pm$ 0.024 & 0.844 $\pm$ 0.015 & 0.906 $\pm$ 0.005 & 0.568 $\pm$ 0.002 & 16.2\\
	\cmidrule{2-13}

 &   \multicolumn{5}{c}{{\textbf{\underline{Decoder Convolutions}}}}\\

  & \multicolumn{3}{r}{Remaining convs} & \multicolumn{2}{l}{Descriptor conv}\\
   &  & \scriptsize{Int8} & \scriptsize{BIN-R} & \scriptsize{FP} & \scriptsize{Int8}\\
&\hspace*{1cm}\multirow{2}{*}{\reflectbox{$\WithArrowsOptions{displaystyle,tikz=ForestGreen}\begin{WithArrows} & \Arrow{}[jump=2] \\&  \\&  \\& \\& \\& \end{WithArrows}$}\vspace{-1.2cm}}\hspace*{-0.7cm}	&\textcolor{ForestGreen}{\pmb{\checkmark}}&&\textcolor{ForestGreen}{\pmb{\checkmark}}& & 0.658 $\pm$ 0.002 & 0.964 $\pm$ 0.020 & 0.488 $\pm$ 0.030 & 0.840 $\pm$ 0.013 & 0.900 $\pm$ 0.001 & 0.569 $\pm$ 0.002 & 27.2\\
	&&&\pmb{\checkmark}&\pmb{\checkmark}& & 0.655 $\pm$ 0.003 & 1.018 $\pm$ 0.006 & 0.451 $\pm$ 0.001 & 0.456 $\pm$ 0.001 & 0.481 $\pm$ 0.002 & 0.329 $\pm$ 0.008 &30.1\\
		\gcmidrule{2-13}
	&&\pmb{\checkmark}&&& \pmb{\checkmark} & \textbf{0.652 $\pm$ 0.005} & \textbf{0.926 $\pm$ 0.022} & \textbf{0.506 $\pm$ 0.025} & \textbf{0.853 $\pm$ 0.007} & \textbf{0.917 $\pm$ 0.003} & \textbf{0.571 $\pm$ 0.003} &
	\textbf{47.2}\\
    \bottomrule
  \end{tabularx}}
  \vspace{-0.2in}
\end{table*}

%% file: text/exp.tex
\section{Experiments}
\label{sec:exp}

We present the implementation details in Sec.~\ref{sec:impl_details}. 
In Sec.~\ref{sec:quant_ablation} we investigate the effect of network quantization using the layer partitioning and traversal strategy, as well as evaluate the proposed \gls{Bin.Norm} layer for descriptor binarization.
We then combine the two contributions into \OURS{} and evaluate its performance on the task of homography estimation.
We evaluate the generalization capabilities of \OURS{} on fundamental tasks in robotic and \gls{AR} pipelines, namely \gls{VisLoc} in Sec.~\ref{sec:vis_loc}, and Map-Free Visual Relocalization in 
% the supplementary material.
Sec.~\ref{sec:mapfreereloc}.
We envision our work can both spark further research in the design of binary descriptors and quantized networks, as well as promote the incorporation of \OURS{} in robotic systems.

\subsection{Implementation Details}
\label{sec:impl_details}

We implement our models in TensorFlow~\cite{tensorflow2015-whitepaper}, and use the Larq~\cite{larq} library for quantization.
Our models are trained on the COCO 2017 dataset~\cite{lin2014microsoft}, comprised of 118k training images, following ~\cite{detone2018superpoint,christiansen2019unsuperpoint,tang2019neural}.
The models are optimized using ADAM~\cite{kingma2014adam} for 50 epochs with a batch size of 8, starting with an initial learning rate of $10^{-3}$ while halving it every 10 epochs.
To ensure robustness in our results, we optimize each model configuration three times and report the mean and standard deviation.

To enable self-supervised training, spatial and non-spatial augmentations for the homography transformation are required.
For spatial transformations, we utilize crop, translation, scale, rotation, and symmetric perspective.
Non-spatial transformations applied are per-pixel Gaussian noise, Gaussian blur, color augmentation in brightness, contrast, saturation, and hue.
Finally, we randomly shuffle the color channels and convert images to gray scale.
Please refer to \cite{tang2019neural} for more details.

\subsection{Designing \OURS{}}
\label{sec:quant_ablation}

We conduct our DNN quantization investigation on the task of homography estimation, a commonly used task for the evaluation of self-supervised learned models~\cite{detone2018superpoint,christiansen2019unsuperpoint,tang2019neural}.
Homographic transformations largely eliminates domain shifts due to missing 3D, providing a good benchmark for ablation studies.

We evaluate our method on image sequences from the HPatches dataset~\cite{balntas2017hpatches}.
HPatches contains 116 scenes, separated in 57 illumination and 59 viewpoint sequences.
Each sequence is comprised of 6 images, with the first image used as a reference.
The remaining images are used to form pairs for evaluation.
As is common practice, we report Repeatability (Repeat.), Localization Error (Loc), Matching Score (M.Score), and Homography Accuracy with thresholds of 1, 3 and 5 pixels (Cor-1, Cor-3, Cor-5). 
We additionally benchmark and report the CPU speeds in \gls{FPS} on an Apple M1 ARM processor.

\parsection{Baseline.}
We initiate our investigation in Table~\ref{table:layer_part} from a re-implementation of KP2D with minor modifications to enable a structured search with minimal macro-block interference.
Specifically, KP2D uses a shortcut connection between the encoder and decoder macro-blocks.
We remove this skip-connection to constrain the interaction between two macro-blocks to a single point.
Furthermore, we replace the leaky ReLUs with hard-swish~\cite{howard2019searching}, a comparable but faster alternative.
The functional performance of \textit{Baseline} is comparable to KP2D while slightly improving the throughput.

We then partition our baseline architecture into macro-blocks. These include the first encoder convolution, the remaining encoder convolutions, spatial reduction layers, the non-head decoder convolutions, and the head decoder convolutions, as depicted in Fig.~\ref{fig:architecture} by the different colours.

\parsection{Macro-Block I: First Encoder Convolution.}
For macro-block~I, we considered two configurations: FP and Int8.
Although \cite{rastegari2016xnor,liu2018bi} suggest that keeping the first convolution in FP has a negligible effect on application throughput while degradation of functional performance, our findings suggest otherwise.
Specifically, we find that using an Int8 convolution improves throughput by as much as $3$~\gls{FPS}, while having no detectable impact on functional performance.
We ascribe this to the fact that the input images are also represented in Int8.
Therefore, discretization of the input sequence does not cause a loss of information, while enabling the use of a more efficient Int8 convolution.

\begin{table*}[ht]
  \center
    \caption{We evaluate the efficacy of different normalization layers when combined with using sigmoid as a soft approximation for every bit. We find that the binary normalization (Bin.Norm) layer for the descriptors consistently improves all metrics.}
    \vspace{-0.1in}
    \label{table:bin_desc}
      \resizebox{\linewidth}{!}{%
      % \scalebox{0.9}{
      \scriptsize
  \begin{tabularx}{\linewidth}{lcXXXXXr}
    \toprule
    & Norm& Repeat. $\uparrow$ & Loc. $\downarrow$ & Cor-1 $\uparrow$ & Cor-3 $\uparrow$ & Cor-5 $\uparrow$ & M.Score $\uparrow$ \\
    \cmidrule{3-8}
    \rowcolor{Gray}
    \textbf{Full Precision} & $L_2$ & 0.644  $\pm$ 0.003 & 0.788  $\pm$ 0.005 & 0.580  $\pm$ 0.007 & 0.886  $\pm$ 0.008 & 0.933  $\pm$ 0.010 & 0.569  $\pm$ 0.003\\
    \midrule
    \textbf{Sigmoid}  & & 0.640 $\pm$ 0.005 & 0.809 $\pm$ 0.049 & 0.173 $\pm$ 0.300 & 0.285 $\pm$ 0.493 & 0.305 $\pm$ 0.528 & 0.187 $\pm$ 0.318\\
    
    \textbf{Sigmoid + $||L_2||$} & $L_2$ & 0.650 $\pm$ 0.001 & 0.803 $\pm$ 0.010 & 0.491 $\pm$ 0.015 & 0.822 $\pm$ 0.009 & 0.888 $\pm$ 0.003 & 0.513 $\pm$ 0.003 \\
    
    \textbf{Sigmoid + Bin.Norm (Ours)}& Bin.Norm & \textbf{0.651 $\pm$ 0.003} & \textbf{0.796 $\pm$ 0.016} & \textbf{0.545 $\pm$ 0.005} & \textbf{0.880 $\pm$ 0.090} & \textbf{0.925 $\pm$ 0.004} & \textbf{0.553 $\pm$ 0.002}\\
    \bottomrule
  \end{tabularx}}
  \vspace{-0.1in}
\end{table*}

\begin{table*}[ht!]
  \center
    \caption{We compare \OURS{} with full precision or binary descriptors against state-of-the-art methods. \OURS{} performs on par with other full-precision methods while running an order of magnitude faster than the full-precision alternative. When compared to binary hand-crafted methods, \OURS{} consistently outperforms all other methods, often by a large margin.}
    \vspace{-0.1in}
    \label{table:sota_descriptors}
      \resizebox{\linewidth}{!}{%
      % \scalebox{0.9}{
      \footnotesize
  \begin{tabularx}{\linewidth}{lcXXXXXr}
    \toprule
    && Repeat. $\uparrow$ & Loc. $\downarrow$ & Cor-1 $\uparrow$ & Cor-3 $\uparrow$ & Cor-5 $\uparrow$ & M.Score $\uparrow$ \\
    \cmidrule{3-8}
    \rowcolor{Gray}
    \multicolumn{8}{c}{\textbf{\underline{Full-Precision Descriptors}}}\\
    \rowcolor{Gray}
	SuperPoint~\cite{detone2018superpoint} && 0.631 & 1.109 & 0.491 & 0.833 & 0.893 & 0.318\\% & 11.5
	\rowcolor{Gray}
	SIFT~\cite{lowe2004distinctive} && 0.451 & \textbf{0.855} & \textbf{0.622} & 0.845 & 0.878 & 0.304\\
	\rowcolor{Gray}
	SURF~\cite{bay2006surf} && 0.491 & 1.150 & 0.397 & 0.702 & 0.762 & 0.255\\
	\rowcolor{Gray}
	KP2D~\cite{tang2019neural} && \textbf{0.686} & 0.890 & 0.591 & \textbf{0.867} & 0.912 & 0.544\\% & 9.5 \\
	\rowcolor{Gray}
	\OURS{} (Ours) && 0.652 $\pm$ 0.005 & 0.926 $\pm$ 0.022 & 0.506 $\pm$ 0.025 & 0.853 $\pm$ 0.007 & \textbf{0.917 $\pm$ 0.003} & \textbf{0.571 $\pm$ 0.003}\\% & \textbf{47.2} 
	\midrule
    \multicolumn{8}{c}{\multirow{1}{*}{\textbf{\underline{Binary Descriptors}}}}\\
    BRISK~\cite{leutenegger2011brisk} && 0.566 & 1.077 & 0.414 & 0.767 & 0.826 & 0.258\\% & --\\
    ORB~\cite{rublee2011orb} && 0.532 & 1.429 & 0.131 & 0.422 & 0.540 & 0.218\\% & 401.9
    \OURS{} (Ours) && \textbf{0.652 $\pm$ 0.005} & \textbf{0.926 $\pm$ 0.022} & \textbf{0.433 $\pm$ 0.007} & \textbf{0.820 $\pm$ 0.007} & \textbf{0.887 $\pm$ 0.006} & \textbf{0.571 $\pm$ 0.003} \\
    \bottomrule
  \end{tabularx}}
  \vspace{-0.2in}
\end{table*}

\parsection{Macro-Block II: Encoder Convolutions.}
For the encoder convolutions, we considered three configurations: Int8, \gls{Bin}, and \gls{Bin-R}.
While using \gls{Bin} convolutions in the encoder significantly improves application throughput, functional performance is severely hindered, as measured by the halving of the correctness metrics.
This drop is consistent with findings in the literature for semantic segmentation~\cite{zhuang2019structured} while conflicting with image-level classification experiments~\cite{rastegari2016xnor}.
This further supports our arguments for the importance of task-specific investigations.

To alleviate such drastic performance drops, we introduce high precision Int8 representations in the form of a residual operation.
For convolutional operations with a mismatch in the number of input and output channels, we introduce additional Int8 $1 {\times} 1$ convolutions on the residual path.
This ensures the high-precision paths maintain their Int8 precision, while matching the channel dimensions.
The additional high-precision Int8 residuals improve performance significantly.
This again advocates for the redundancy of \gls{FP} representation in the encoder, as the encoder is now bottlenecked by Int8 precision. 

\parsection{Macro-Block III: Spatial Reduction.}
For the spatial reduction layers, we considered four configurations: max-pooling (Max), average-pooling (Aver.), sub-sampling (Sub.S.) and a learned projection (Learn).
As is common in \gls{DNNs}, our baseline utilizes max-pooling. 
However, max-pooling has been found to favour saturated regimes and therefore eliminates information when applied on low-precision features like those found in \gls{QNN}s~\cite{rastegari2016xnor}.
Average pooling further degrades the performance,  attributed to the errors introduced due to the roundings and truncations which are essential for integerized arrays.
To further highlight this error, a simple Sub.S. that only uses information from a quarter of the kernel window yields comparable performance to Aver.

To alleviate the challenges highlighted above, we propose the use of a learned pooling operation (Learn). 
The learned pooling comes in the form of an Int8 convolutional operation with the same kernel size and stride as the other pooling operations.
We select Int8 so as to maintain the representational precision of the network, defined by the macro-block~I and the high precision residuals.
While operating at a comparable run-time to max pooling, the performance significantly improved.
This further corroborates our hypothesis that learned pooling can address both the aforementioned challenges.
Finally, we investigate the effect of the pooling placement (E.Learn).
Specifically, we change the location of the pooling operations from the end to the beginning of each convolutional block.
While with \gls{FP} convolutional layers this would cause a $4\times$ speedup for each convolution, in quantized convolutions the gain is even greater~\cite{larq}.

\vspace{0.1in}

\parsection{Macro-Block IV: Decoder Convolutions.}
For the decoder convolutions, we considered two configurations: Int8, and \gls{Bin-R}.
We do not investigate \gls{Bin} due to the large performance drop observed in macro-block~II.
Unlike the findings from macro-block~II, our decoder experiments demonstrate the importance of Int8, highlighting the benefits of \gls{MP} networks.
Specifically, utilizing Int8 for the entire network would not yield the best throughput, as seen in macro-block~II, while \gls{Bin-R} for the entire network would not yield the best performance.
 
\parsection{Macro-Block V: Final Decoder Convolutions.}
For the final convolutions, we evaluated \gls{FP} and Int8 for the score, location and descriptor heads independently.
We find that score and location heads require \gls{FP} representations, with models often failing to optimize otherwise. 
On the other hand, the descriptor branch can be optimized with Int8, significantly improving the throughput.

\parsection{Network Quantization Findings.}
Network latency can be significantly improved when quantizing the first convolutional layer and the last descriptor head to Int8, while having an insignificant effect on functional performance. 
This is contrary to findings from prior works~\cite{rastegari2016xnor,liu2018bi}.
In addition, enhanced performance can be achieved through \gls{MP} \gls{QNN}s.
In other words, a network comprised of only \gls{Bin-R} or Int8 convolutional operations would yield sub-optimal results.
This observation suggests that good quality and general-purpose features can be extracted using low-precision convolutions when coupled with higher precision residuals.
Furthermore, it suggests that the dense task predictor heads benefit from higher Int8 precision to accurately reconstruct the target information from the encoded features.
Finally, we observe that the prediction heads for regression tasks (score and location) cannot be quantized and should be left in \gls{FP}, while the descriptor head can be quantized to Int8.
This further drives the importance of the structured investigation, like the layer partitioning and traversal strategy.

\parsection{Binarizing descriptors.}
We initiate the descriptor exploration from the common practice of utilizing sigmoid as a soft approximation for every bit~\cite{lai2015simultaneous,liu2012supervised}, and the hamming triplet loss proposed by \cite{lai2015simultaneous}.
While some works use a hard sign function~\cite{tang2019gcnv2}, we found it unable to optimizate the network to a meaningful degree.
Table~\ref{table:bin_desc} demonstrates a significant performance drop and large variance compared to the baseline, especially in the correctness metrics.
We conjecture that this spans from the lack of a normalization layer, causing the sigmoid to saturate, and yielding uninformative gradients.
To test this assumption, we append an $||L_2||$ normalization layer after the element-wise sigmoid operation.
This constrains the activations and dramatically improves performance and reduces the variance, as seen experimentally, leading to a more stable optimization process.

In this paper we hypothesize that, analogous to $||L_2||$ normalization, we can optimize the network using a \gls{Bin} normalization layer by constraining the descriptor to a constant number of ones. 
Using the proposed \gls{Bin.Norm} layer, the functional performance gap is significantly decreased when compared to the \gls{FP} descriptors. 

\parsection{Comparison to state-of-the-art.}
We compare \OURS{} with state-of-the-art methods in Table~\ref{table:sota_descriptors}.
For a fair comparison, we group methods given the descriptor precision.
When utilizing \gls{FP} descriptors, \OURS{} performs on par with other methods.
In particular, it consistently outperforms SuperPoint and performs on par with KP2D.
Meanwhile, the throughput gain is higher than an order of magnitude.

The benefits of \OURS{}, the combination of the fast \gls{QNN} architecture from Table~\ref{table:layer_part} and the binary optimization strategy from Table~\ref{table:bin_desc}, become apparent when comparing binary descriptor methods. 
We consistently outperform ORB~\cite{mur2015orb} by a large margin in all metrics.
We additionally outperform BRISK~\cite{leutenegger2011brisk} in all metrics and even report double the matching score, a crucial metric for adaptation of these methods in downstream tasks like \gls{VisLoc}.
Here on, we refer to \OURS{} as our \gls{QNN} with binary descriptors.

\begin{figure}[t]
 \centering
\includegraphics[width=\columnwidth]{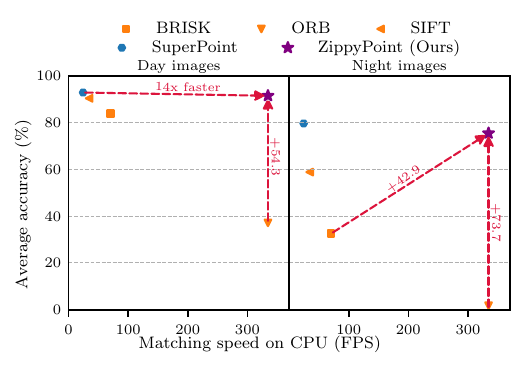}
\caption{Comparison of the average visual localization accuracy vs descriptor matching speed between two images on the AachenV1.1 Day-Night datasets. \OURS{} consistently outperforms all other binary methods.}\label{fig:aachen_loc}
\vspace{-0.2in}
\end{figure}

\subsection{Visual Localization}
\label{sec:vis_loc}
Camera localization is one of the key components in several robotic and mapping applications.
Both relative~\cite{nister2004efficient} and absolute~\cite{kneip2011novel} camera localization require good local feature point descriptors to match, and are key building blocks in seminal  pipelines~\cite{mur2015orb,mur2017orb,endres2012evaluation,galvez2012bags,leutenegger2015keyframe,davison2007monoslam}.
To further demonstrate the potential of \OURS{}, we assess its generalization capability on the task of absolute camera localization, where the pose of a query image is estimated with respect to a 3D map.

We utilize the hloc framework~\cite{sarlin2019coarse}, similar to prior works \cite{revaud2019r2d2,sarlin2020superglue}, and evaluate the performance on the challenging real-life AachenV1.1 Day-Night datasets from the \gls{VisLoc} benchmark~\cite{sattler2018benchmarking,sattler2012image}. 
More precisely, we reconstruct the 3D map using \OURS{} features instead of SIFT~\cite{lowe2004distinctive}.
For each query image, we perform a coarse search of the map and retrieve the 30 closest database images based on their global descriptors, representing candidate locations.
The query image is then localized within the 3D map by utilizing the candidate locations.
Please refer to \cite{sarlin2019coarse} for more details. 

The results are presented in Fig.~\ref{fig:aachen_loc} with respect to the \gls{FPS} speed for matching two images.
In Fig.~\ref{fig:teaser} we additionally depict the average performance score for both day and night query sets with respect to 3D model size, query localization time, and model inference speed.
While \OURS{} performs comparably to SuperPoint during day time, we decrease the 3D model size, query localization time, and model inference speed by at least an order of magnitude. 
This is attributed to the lightweight binary descriptors, the more efficient similarity comparison between the descriptors, and the network quantization.
Localization with \OURS{} at night is slightly inferior to SuperPoint, however, we expect optimization of the image transformations during training can close this gap further.

On the binary descriptor front, \OURS{} consistently outperforms ORB by a significant margin at a comparable matching speed.
BRISK on the other hand is competitive to ours on the day dataset, with the slower run-time of BRISK attributed partly to the larger descriptor size, twice that of \OURS{}, and the increased number of detected keypoints.
However, the more challenging night dataset paints a different picture, with \OURS{} outperforming BRISK by 42.9\% and ORB failing to localize. 
This further attests to the need for efficient learned detection and description networks, in particular for more challenging and adverse conditions.

%% file: text/conclusion.tex
\section{Conclusion}
\label{sec:conclusion}
In this paper, we investigated efficient detection and description of learned local image points through mixed-precision quantization of network components and binarization of descriptors. 
To that end, we followed a structured investigation, we refer to as layer partitioning and traversal for the quantization of the network.
In addition, we proposed the use of a binary normalization layer to generate binary descriptors with a constant number of ones.

\renewcommand\thefigure{S.\arabic{figure}}    
\renewcommand\thetable{S.\arabic{table}}   
\appendix

\begin{table*}[!b]
  \center
    \caption{Comparison of the visual localization accuracy, given different error threshold, on the AachenV1.1 Day-Night datasets. We additionally report the 3D model size (Map), the localization speed (Loc.) for descriptor extraction and matching in the hloc framework~\cite{sarlin2019coarse}, the inference speed for the extraction of the descriptors (Inf.), and the matching speed for two images (Match.).
    The arrows indicate the improvement direction.
    \OURS{} consistently outperforms all other binary descriptor methods, while yielding great trade-offs with respect to inference speed, matching speed, and model size.}
    \label{sup_table:aachen_loc}
      \resizebox{\linewidth}{!}{%
      % \scalebox{0.9}{
      \footnotesize
  % \begin{tabularx}{\linewidth}{lcXXXXXr}
  \begin{tabularx}{0.95\linewidth}{lcccc|ccc|cccc}
    \toprule
    && \multicolumn{3}{c}{Day $\uparrow$} & \multicolumn{3}{c}{Night $\uparrow$} & Map (MB) $\downarrow$ & Loc. (FPS) $\uparrow$  & Inf. (FPS) $\uparrow$ & Match. (FPS) $\uparrow$  \\
    &m& 0.25 & 0.50 & 5.00 & 0.50 & 1.00 & 5.00 \\
    &deg& 2 & 5 & 10 &2 & 5 & 10 \\
    \cmidrule{3-12}
    \rowcolor{Gray}
    \multicolumn{12}{c}{\textbf{\underline{Full-Precision Descriptors}}}\\
    \rowcolor{Gray}
	SuperPoint~\cite{detone2018superpoint} && 
 86.8 & 93.8 & 97.9 & 
 62.3 & 81.7 & 94.8 & 
 5224 & 
 0.22 & 
  0.29 & 
 24.4 \\
	\rowcolor{Gray}
	SIFT~\cite{lowe2004distinctive} && 
 82.3 & 91.6 & 97.0 & 
 45.0 & 58.6 & 72.8 & 
 3756 & 
  1.00 & 
  7.93 & 
  34.5 \\
	\midrule
	\multicolumn{12}{c}{\textbf{\underline{Binary Descriptors}}}\\
    BRISK~\cite{leutenegger2011brisk} && 
    75.2 & 84.1 & 92.4 & 
    23.0 & 32.5 & 41.9 & 
    638 & 
   1.11 & 
  2.10 & 
  70.4\\
    ORB~\cite{rublee2011orb} && 
    25.4 & 35.3 & 50.6 & 
    \phantom{0}1.0 & \phantom{0}1.6 & \phantom{0}2.6 & 
    113 & 
  10.39& 
  54.80& 
  334.5\\
    \OURS{} (Ours) && 
    \textbf{85.0} & \textbf{92.2} & \textbf{97.0} &
    \textbf{63.4} & \textbf{74.9} & \textbf{88.0} &
    163 & 
  3.47 & 
  4.76 & 
  334.5\\
    \bottomrule
  \end{tabularx}}
\end{table*}

We obtained an order of magnitude throughput improvement with minor degradation of performance.
In addition, we find that the binary normalization layer allows the network to operate on par with full-precision networks, while consistently outperforming hand-crafted binary descriptor methods.
The results show the suitability of our approach on visual localization and map-free visual relocalization, challenging downstream tasks and essential prerequisites for robotic applications, while significantly decreasing the 3D model size, matching, and localization speed.
We believe \OURS{} can spark further research towards bringing learned binary descriptor methods to mobile platforms, as well as promote its incorporation in both new and established robotic pipelines.

%% file: text_supp/visloc_additional.tex
\section{Visual Localization}
\label{sec:vis_loc_sup}
% Fig.~\textcolor{red}{3} from the main paper 
Fig.~\ref{fig:aachen_loc}
presents the average visual localization accuracy of the AachenV1.1 Day-Night datasets~\cite{sattler2012image,sattler2018benchmarking}, with respect to the descriptor matching speed between two images.
Table~\ref{sup_table:aachen_loc} presents the performance breakdown for both the day and night datasets. 
In addition, we report the 3D map size (Map) in megabytes, the localization speed (Loc.) for descriptor extraction and matching in the hloc framework~\cite{sarlin2019coarse}, the inference speed for the extraction of the descriptors (Inf.), and the matching speed for two images (Match.).
All speeds are calculated on an Apple M1 ARM CPU processor and are reported in \gls{FPS}.
As seen, \OURS{} consistently outperforms all other binary descriptor methods, while yielding great trade-offs with respect to inference speed, matching speed, localization speed, and model size.
Note that, the inference speed reported in Table~\ref{sup_table:aachen_loc} is lower than that of 
% Table.~\textcolor{red}{1} from the main paper.
Table~\ref{table:layer_part}.
This is attributed to the fact that the inference speed for learned methods scales linearly with the number of spatial dimensions in the input image. 
The image resolution used in 
% Table.~\textcolor{red}{1} 
Table~\ref{table:layer_part}
was $240 \times 320$, following \cite{tang2019neural}, while in Table~\ref{sup_table:aachen_loc} the largest image dimension was rescaled to $1020$, following~\cite{sarlin2019coarse}.

%% file: text_supp/mapfreeloc.tex
\begin{table*}[t]
  \center
    \caption{Comparison of the different detection and description networks on the Map-free Visual Relocalization benchmark~\cite{arnold2022map} at the original image resolution.
We report the Area Under the Curve (AUC) and precision  under the Virtual Correspondence Reprojection Error (VCRE) and pose error (Err) with respect to the feature extraction and image matching speed (Latency) in seconds (s). 
    \OURS{} yields comparable performance to SuperPoint while being an order of magnitude faster.
Additionally, \OURS{} consistently outperforms the binary methods, BRISK and ORB, by a large margin.}
    \label{sup_table:mapfreereloc}
      % \resizebox{\linewidth}{!}{%
      % \scalebox{0.9}{
      \footnotesize
  % \begin{tabularx}{\linewidth}{lcXXXXXr}
  \begin{tabularx}{0.8\linewidth}{lcccccc}
    \toprule
    && \multicolumn{2}{c}{AUC $\uparrow$} & \multicolumn{2}{c}{Precision $\uparrow$} & Latency (s)  $\downarrow$  \\

    && VCRE $<$ 90px & Err $<$ 25cm, 5$\deg$ & VCRE $<$ 90px & Err $<$ 25cm, 5$\deg$ &\\
    \cmidrule{3-7}
    \rowcolor{Gray}
    \multicolumn{7}{c}{\textbf{\underline{Full-Precision Descriptors}}}\\
    \rowcolor{Gray}
	SuperPoint~\cite{detone2018superpoint} && 
  0.405& 
  0.199& 
  0.231& 
  0.090& 
  1.678\\
	\rowcolor{Gray}
	SIFT~\cite{lowe2004distinctive} && 
 0.443& 
  0.189 & 
  0.217& 
  0.076& 
  0.068\\
	\midrule
	\multicolumn{7}{c}{\textbf{\underline{Binary Descriptors}}}\\
    BRISK~\cite{leutenegger2011brisk} && 
   0.307 & 
  0.120& 
  0.181& 
  0.054& 
  0.304\\
    ORB~\cite{rublee2011orb} && 
   0.044 & 
0.013& 
  0.033& 
  0.007& 
  0.009\\
    \OURS{} (Ours) && 
   0.415 & 
  0.206& 
  0.192& 
  0.074& 
  0.107\\
    \bottomrule
  \end{tabularx}
\end{table*}

\begin{figure*}[t]
 \centering
    \subfloat[Virtual Correspondence
Reprojection Error AUC.\label{fig:AUC_VCRE}]{\includegraphics[width=0.5\linewidth]{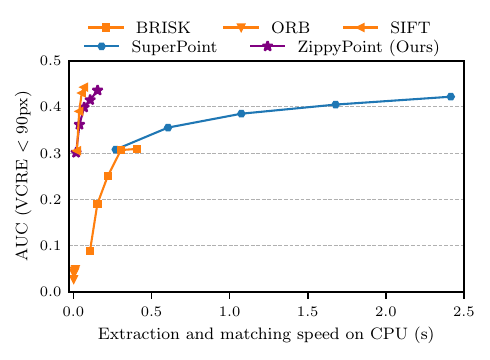}}
    ~
    \subfloat[Virtual Correspondence
Reprojection Error precision.\label{fig:Precision_VCRE}]{\includegraphics[width=0.5\linewidth]{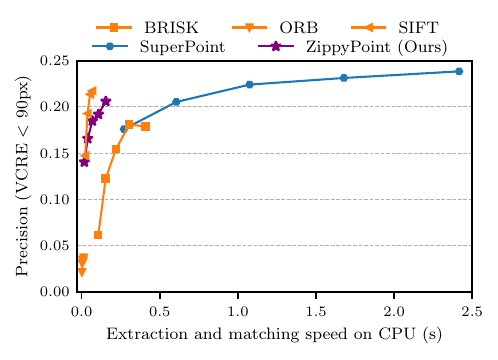}}

    \subfloat[Pose error AUC.\label{fig:AUC_Pose}]{\includegraphics[width=0.5\linewidth]{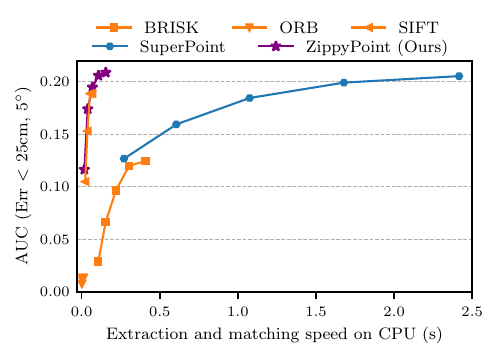}}
    ~
    \subfloat[Pose error precision.\label{fig:Precision_Pose}]{\includegraphics[width=0.5\linewidth]{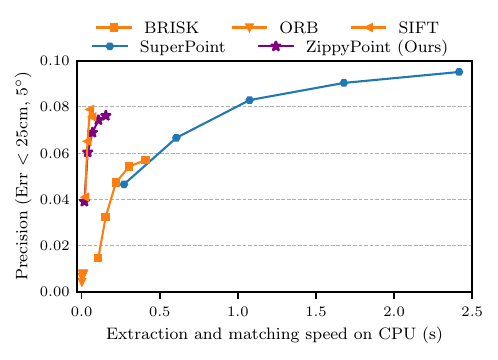}}
\caption{Comparison of the different detection and description networks on the Map-free Visual Relocalization benchmark~\cite{arnold2022map}.
We report the Area Under the Curve (AUC) and precision  under the Virtual Correspondence Reprojection Error (VCRE) and pose error (Err) with respect to the feature extraction and image matching speed. 
\OURS{} consistently outperforms all binary descriptor methods and achieves comparable performance to SuperPoint at a significant speedup.} 
\label{fig:mapfreereloc}
% \vspace{2.3in}
\end{figure*}

\section{Map-free Visual Relocalization}
\label{sec:mapfreereloc}

Absolute camera localization, such as the task presented in 
Sec.~\ref{sec:vis_loc} 
% Sec.~\textcolor{red}{4.3}
and Sec.~\ref{sec:vis_loc_sup}, require an accurate 3D scene-specific map.
This entails hundreds of images and large storage space, prerequisites that do not often hold in \gls{AR} applications.
These limitations have given rise to the more challenging Map-free Visual Relocalization benchmark~\cite{arnold2022map}.
The aim of Map-free Visual Relocalization is to predict the metric pose of a query image with respect to a single reference image that is considered representative of the scene of interest.

We evaluate interest point detection and description networks on the challenging Map-free Visual Relocalization benchmark.
Specifically, as in \cite{arnold2022map}, we first 
compute the Essential matrix~\cite{hartley2003multiple} between the query and the reference image using the 5-point solver~\cite{nister2004efficient} of \cite{barath2020magsac++}. 
We then recover the scale using the estimated depth generated from a DPT model~\cite{ranftl2021vision} that has been fine-tuned on the KITTI dataset~\cite{geiger2012we}.
We report the \gls{AUC} and precision for \gls{Err} under the threshold of 25cm and 5-degree.
In addition, we report \gls{AUC} and \gls{Err} for  \gls{VCRE} at an offset threshold of 10\%, 90 pixels, simulating the placement of \gls{AR} content in the scene~\cite{arnold2022map}.
The performances are reported in Fig.~\ref{fig:mapfreereloc} and Table~\ref{sup_table:mapfreereloc} with respect to the latency for keypoint extraction and matching.
For Fig.~\ref{fig:mapfreereloc}, we identify Pareto curves by rescaling the input images at ratios of 0.4 to 1.0 in 0.2 increments, a common practice to accelerate inference post-training, and also increase the ratio to 1.2 in order to evaluate if performance can improve further, as commonly done in \gls{VisLoc} \cite{sarlin2019coarse}.
We also investigated larger ratios but found they often degraded the performance of the hand-crafted methods, such as SIFT, while the performance quickly plateaued for the \gls{DNNs}.

We find that \OURS{} yields comparable performance to SuperPoint while being an order of magnitude faster for feature extraction and matching.
Additionally, \OURS{} consistently outperforms the binary methods, BRISK and ORB, by a large margin.
When compared to SIFT~\cite{lowe2004distinctive}, however, \OURS{} yields comparable results at a slight increase in latency.
This is attributed to the nature of the dataset and task.
Specifically, the Map-free Visual Relocalization benchmark presents a wide baseline benchmark without challenging long-term changes, the scenario under which SIFT shines.
We expect similar benchmarks with long-term changes, similar to \gls{VisLoc}, would better showcase the benefits of \OURS{}, and the learned methods in general.
Furthermore, while SIFT's keypoint matching is slower than ZippyPoint's, matching only takes place between a single pair of images for each scene in this experiment and therefore does not aggregate to a significantly large delay, unlike in \gls{VisLoc} and \gls{SLAM} where matching speed is often the bottleneck due to the required matching within a large map.